\pdfoutput=1

\documentclass[11pt]{article}

\usepackage[]{coling}

\usepackage{times}
\usepackage{latexsym}

\usepackage[T1]{fontenc}

\usepackage[utf8]{inputenc}

\usepackage{microtype}

\usepackage{inconsolata}
\usepackage{graphicx} 
\usepackage{algorithm}
\usepackage{algorithmic}
\usepackage{amsmath} 
\usepackage{amsfonts}
\usepackage{newfloat}
\usepackage{listings}
\usepackage{multirow}
\usepackage{makecell}
\usepackage{subfig}
\usepackage{dsfont} 
\usepackage[most]{tcolorbox}
\usepackage{hyperref}
\usepackage{xcolor}
\usepackage{arydshln}
\definecolor{PastelPurple}{HTML}{b0a4e3}
\title{\hspace{-0.5cm}\raisebox{-0.45cm}{\includegraphics[width=1.5cm,height=1.5cm,keepaspectratio]{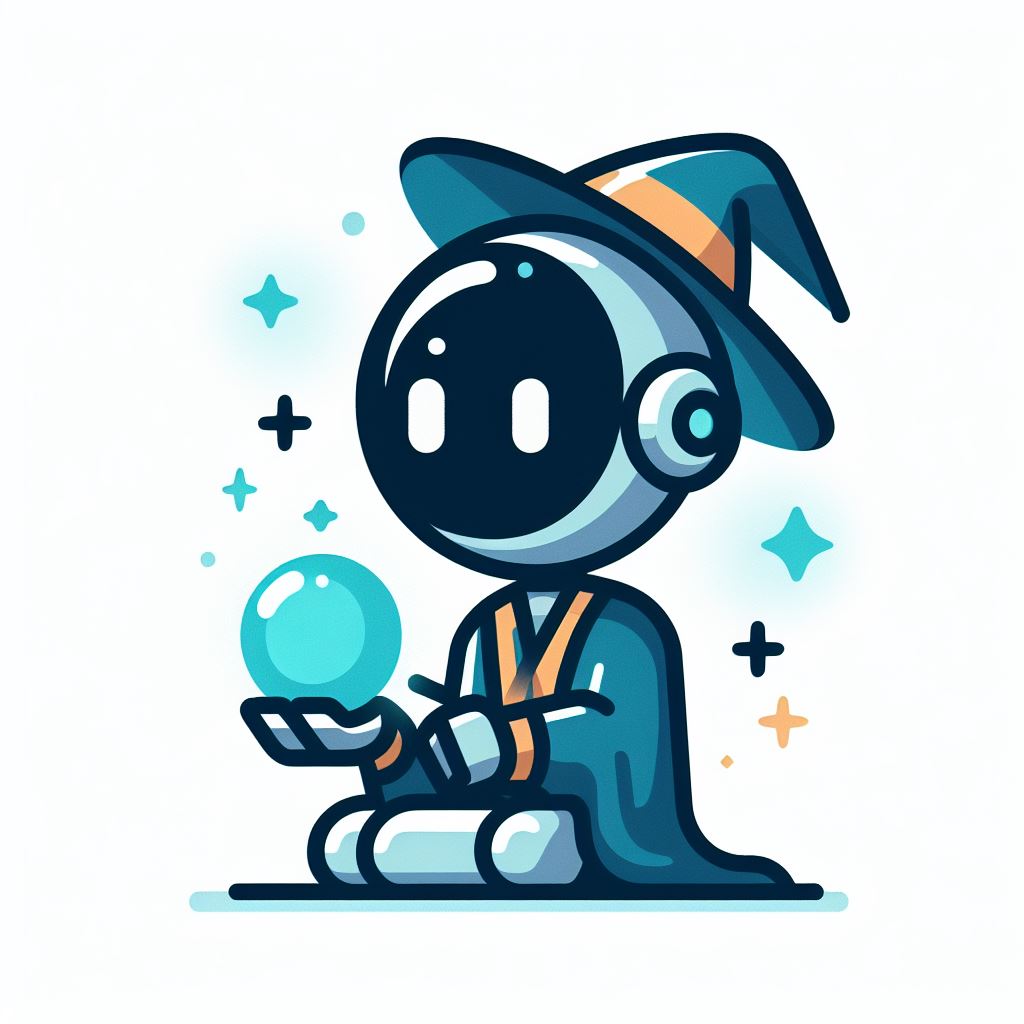}}PreAct: Prediction Enhances Agent's Planning Ability}

\author{Dayuan Fu$^{1}$, Jianzhao Huang$^{1}$, Siyuan Lu$^{1}$, Guanting Dong $^{1}$, \\ {\bf Yejie Wang$^{1}$},    {\bf Keqing He$^{2}$},  {\bf Weiran Xu$^{1}$}\thanks{\ \ Weiran Xu is the corresponding author.}\\
  $^1$Beijing University of Posts and Telecommunications, Beijing, China\\
$^{2}$Meituan, Beijing, China\\
  \texttt{\{fdy,huangjianzhao,lengshuiyu,dongguanting,wangyejie,xuweiran\}@bupt.edu.cn}\\
  \texttt{hekeqing@meituan.com}
  }

\begin{document}
\maketitle
\begin{abstract}

Addressing the disparity between forecasts and actual results can enable individuals to expand their thought processes and stimulate self-reflection, thus promoting accurate planning.
In this research, we present \textbf{PreAct}, an agent framework that integrates \textbf{pre}diction, \textbf{rea}soning, and \textbf{act}ion. By utilizing the information derived from predictions, the large language model (LLM) agent can provide a wider range and more strategically focused reasoning. This leads to more efficient actions that aid the agent in accomplishing intricate tasks. Our experimental results show that PreAct surpasses the ReAct method in completing complex tasks and that PreAct's performance can be further improved when paired with other memory or selection strategy techniques. We presented the model with varying quantities of historical predictions and discovered that these predictions consistently enhance LLM planning.
The variances in single-step reasoning between PreAct and ReAct indicate that PreAct indeed has benefits in terms of diversity and strategic orientation over ReAct. \footnote{Our code will be released at https://github.com/Fu-Dayuan/PreAct.}
\end{abstract}

\section{Introduction}










The language agent is made to address the Markov decision processes (MDPs) issues \citep{wei2022chain,kojima2022large,wang2022self} through the planning and decision-making capabilities of the large language model (LLM) \citep{achiam2023gpt}.
Given that MDPs comprise two main parts, action and state, the optimization of the language agent  can be broken down into 2 questions:

(Q1) Which action(s) should be sampled based on a given state? 

(Q2) Which state is closest to task completion? 

On the one hand, ReAct \citep{yao2022react} found that using chain-of-thought (COT) \citep{wei2022chain} with all historical thought, action, and observation, LLMs can sample higher quality action(s) than Act-only prompt.


On the other hand, TOT \citep{yao2023tree}, GOT \citep{besta2023graph}, and RAP \citep{hao2023reasoning} generate multiple possible actions (ReAct-based or Act-only-based \citep{yao2022react}) and select a state each turn based on state selection strategies and observations. They found good state selection strategies can also improve the overall results. 

Recent work seems to focus mostly on designing superior state selection strategies and overlooking the optimization of action sample methods. 
However, such action sampling methods typically generate direct causal reasoning pathways and may generate the same actions multiple times, which limits their effectiveness in tasks requiring complex relationships. As a result, finding an action sampling method that can improve action quality and actions diversity is important.

\begin{figure}[t]
  \centering
  \includegraphics[width=0.47 \textwidth]{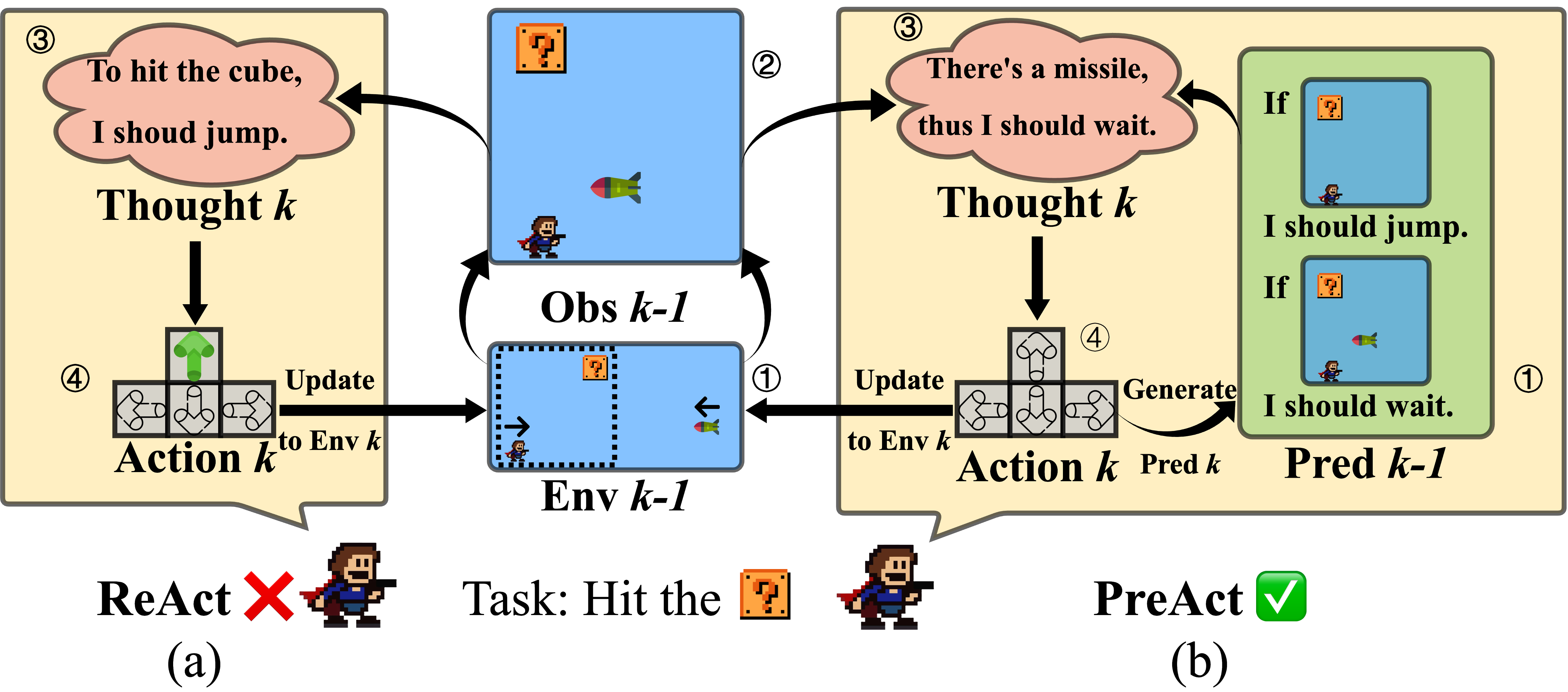}
  \caption{Comparison between ReAct and PreAct. The scene start from $Env_{k-1}$ and $Pred_{k-1}$, The $Obs_{k-1}$ comes from the $Env_{k-1}$ and $Action_{k-1}=go\; right$.
  Env = environment, Obs=observation, Pred=prediction. }
  \label{fig:PreAct framework}
\end{figure}

Inspired by the works in Task-Oriented Dialogue about predicting future \citep{qi2020prophetnet,zeng2022semi,zeng2023futuretod,lei2023instructerc}, we introduce \textbf{PreAct}: \textbf{pre}dict future with \textbf{rea}soning and \textbf{act}ion. Specifically, PreAct requires predicting the possible observations and corresponding measures at a higher level after making an action. This mode can enhance LLMs' directional strategy in reasoning to assist planning in the right way. It can also guide LLMs to conduct more diverse reasoning, thereby leading LLMs to explore thinking more broadly and comprehensively, enabling the agent to handle tasks with greater complexity.

In summary, our main contributions are:

(1) We first propose PreAct, a simple and effective approach to synchronize reasoning, action, and prediction in language models for task-solving.

(2) Our research confirms PreAct's effectiveness, regardless of Reflexion technology or TOT selection strategy. Our experiment demonstrates that PreAct enhances the diversity and directional strategy of planning.

(3) The ablation studies have revealed that predictions play a continuous and positive role in augmenting planning and decision-making.

\section{Method}
\subsection{Preliminaries}
\textbf{Agent in Environment}
Actions and observations construct the process agent made in the environment. For an agent in step $k$, the agent will give an action based on history information, last observation, and its action policy $a_k=\pi_{agent}(o_{k-1}, history)$. After the action has been decided, the agent will act in the environment and gain the new observation by state transition function $o_k=\pi_{env}(o_{k-1}, a_{k})$. For an LLM agent, it can only control the $\pi_{agent}$ and the construction of $history$. So, the target of the LLM agent is to design efficient $\pi_{agent}$ and $history$.

\textbf{ReAct} \citep{yao2022react} ReAct is a pioneering work towards LLM agent that combines thought $t$, action $a$, and observation $o$. ReAct use $LLM( \cdot | COT\; prompt )$ as the $\pi_{agent}(\cdot)$ and the set of $\{o_0,t_1,a_1,o_1,...,t_{k-1},a_{k-1}\}$ as the $history_r$. By leveraging LLM's planning ability, the ReAct agent can explore the environment and solve the problem step by step.

\textbf{Reflexion} \citep{shinn2023reflexion} Reflexion is a long-term memory strategy to improve the quality of $history$ in Agent. Take ReAct's Reflexion as an example, if a task fails, the LLM will be asked to make a reflection like $ref=LLM_{ref}(\{o_0,t_1,a_1,o_1,...,t_{k},a_{k},o_{k}\})$. Once the reflection has been made, the $history$ will be updated to $\{ref,o_0,t_1,a_1,o_1,...,t_{k-1},a_{k-1}\}$. Such a strategy can remind LLM of some information and help it to avoid some decision mistakes.

\textbf{TOT} \citep{yao2023tree} Tree-of-thought is a selection strategy to improve the qualities $actions$. Specifically, TOT will sample several actions and select 1 action in each turn. Its action policy can be formulated as

$\{a_{k1},...a_{kn}\}=\pi_{agent}(o_{k-1}, history)$.

 $a_{k}=\pi_{selection}(o_{k-1}, history, a_{k1},...a_{kn})$

\subsection{PreAct}

 The framework of PreAct has been shown in Figure \ref{fig:PreAct framework}. It has two differences with ReAct. For the $\pi_{agent}(\cdot)$ part, PreAct will prompt the LLM to generate a prediction $p$ of future observation(s) and measurements in each step and hint the LLM to reflect or change its plan direction based on the difference between the predict observation(s) and the real observation. By applying the prompt, the diversity and strategy of the plan LLM made can be enhanced. \footnote{All of the prompt can be found in Appendix \ref{PreAct prompt}.} For the $history$ part, PreAct will add the prediction of future observation(s) in it. \\
Although PreAct seems to improve LLM's reflection and planning ability, there are still 3 questions: 

(1) Do PreAct and Reflexion work orthogonal?

(2) Do PreAct and selection strategies like TOT work in a mutually reinforcing manner?

(3) Is the effect of the prediction permanent? 

Based on these questions, we consider 4 modes:\\ (1) \textbf{Permanent}\footnote{In the following text, the default PreAct mode is the permanent mode and immediate mode will be used in \S 3.4}: All predictions will be preserved in permanent history, as $history_p=\{o_0,t_1,a_1,p_1,o_1,...,t_{k-1},a_{k-1},p_{k-1}\}$ \\(2) \textbf{Immediate}:  Only the last prediction will be preserved in immediate history, as $history_i=\{o_0,t_1,a_1,o_1,...,t_{k-1},a_{k-1},p_{k-1}\}$ \\(3) \textbf{Reflexion}: Reflexion and all prediction will be preserved in the history, as $history_r=\{ref,o_0,t_1,a_1,p_1,o_1,...,t_{k-1},a_{k-1},p_{k-1}\}$ \\(4) \textbf{TOT}: Applying TOT action policy with $history=history_p$ 
 (TOT-PreAct) or $history=history_r$ (TOT-ReAct)

\section{Experiment}

Our experiments are designed to address the following research questions (RQs): \textbf{RQ1}: Does PreAct exhibit higher effectiveness compared to ReAct in dealing with tasks among different modes? 
\textbf{RQ2}: Does historical prediction contribute to sustained gains in planning? \textbf{RQ3}: What are the intrinsic reasons for PreAct's superior facilitation of planning compared to ReAct?

\begin{table}[htbp]
  \centering
    \resizebox{0.45\textwidth}{!}{
    \begin{tabular}{lrrrrrrcc}
\Xhline{1.5pt}
\multicolumn{1}{c}{\multirow{1}[4]{*}{Model}} & \multicolumn{2}{c}{HH} & \multicolumn{2}{c}{OS} & \multicolumn{2}{c}{DB} & \multicolumn{2}{c}{LTP} \\
\cline{2-9}      & \multicolumn{1}{l}{Dev} & \multicolumn{1}{l}{Test} & \multicolumn{1}{l}{Dev} & \multicolumn{1}{l}{Test} & \multicolumn{1}{l}{Dev} & \multicolumn{1}{l}{Test} & \multicolumn{1}{l}{Dev} & \multicolumn{1}{l}{Test} \\
\Xhline{1.5pt}
\textbf{\textit{Permanent Mode}} &       &       &       &       &       &       &       &  \\
ReAct (3.5) & 0.0    & 10.0   & \textbf{46.2}  & 16.7  &\textbf{53.3}  & 39.3  & 13.5 & 11.0 \\
PreAct (3.5) & \textbf{15.0} &\textbf{18.0} &\textbf{46.2} & \textbf{20.1}  & \textbf{53.3}  & \textbf{45.7} & \textbf{16.9} & \textbf{14.1}\\
\hdashline
ReAct (4) & 65.0  & 68.0& 65.4  & 37.5  & 56.7  & 51.3  & 29.7 & \textbf{29.0}\\
PreAct (4) & \textbf{80.0}  &\textbf{78.0}  & \textbf{69.2}  & \textbf{43.1}& \textbf{58.3} & \textbf{51.3} & \textbf{30.6}& 24.9 \\
\hline
\textbf{\textit{Reflexion Mode}} &       &       &       &       &       &       &       &  \\
ReAct (3.5) & 10.0   & 18.0  & 50.0    & 21.5  & 55.0    & 45.6  & -     & - \\
PreAct (3.5) & \textbf{35.0} &\textbf{20.0}   &\textbf{53.8}  & \textbf{24.3}  & \textbf{60.0}    & \textbf{55.3} & -     & - \\
\hdashline
ReAct (4) & 80.0   & 78.0  & \textbf{73.1} & 48.6  & \textbf{61.7}  & 58.0    & -     & - \\
PreAct (4) & \textbf{90.0}   &\textbf{80.0}  & \textbf{73.1}  & \textbf{50.0}   & \textbf{61.7} &\textbf{58.3} & -     & - \\
\Xhline{1.5pt}
\end{tabular}%
    }
  \caption{The result of ReAct and PreAct in 4 datasets. The version of GPT are included in ( parentheses )}
      \label{tab:main result}%
\end{table}%

\begin{table}[tbp]
  \centering
    \resizebox{0.45\textwidth}{!}{
    \begin{tabular}{lcccccc c c}
    \Xhline{1.2pt}
    \multicolumn{1}{c}{\multirow{1}[4]{*}{Model}} & \multicolumn{5}{c}{100} & & \multicolumn{1}{c}{1000} \\
    \cline{2-6} \cline{8-9}    
    & 0 & 1  & 2 & 3 & 4 &  & 0\\
    \Xhline{1.2pt}
    ReAct+TOT & 66 & 63 & 62 & 65 & 67 & &  64.9 \\
    PreAct+TOT & 70  & 72  & 67  & 70 & 68 &  & 70.8 \\
    \Xhline{1.2pt}
    \end{tabular}%
}
  \caption{The result of ReAct and PreAct with TOT in HotpotQA under different sample sizes. We ran the test 5 times with a sample size of 100 and once with a sample size of 1000 due to the inherent randomness of TOT results and the scale of HotpotQA.}
      \label{tab:tot}%
\end{table}%

\subsection{Experiment Setup}

We evaluate PreAct on 4 different sub-datasets, Householding (HH), Operating System (OS), Database (DB), and Lateral thinking puzzles (LTP)\footnote {Due to the distinctiveness of LTP, we only apply LTP in permanent and immediate mode. More details can be found in appendix \ref{ltp info}} in AgentBench \citep{liu2023agentbench}. We evaluate PreAct with TOT on HotpotQA \citep{yang2018hotpotqa}. More details can be found in Appendix \ref{exp-set}.

\subsection{Main Result (RQ1)}
\label{main result}

Table \ref{tab:main result} delineates the performance of PreAct and ReAct under two distinct settings, Permanent and Reflexion, across four datasets. In the HH task, PreAct boasts an approximate 20\% enhancement over ReAct. In the OS and DB coding tasks, there is an average improvement of 12\% and 6\% respectively with PreAct, and under the Reflexion setting, the enhancements are 5\% and 8\% respectively. In the LTP context, PreAct yields results akin to Act-only, which may be attributed to GPT's safety mechanisms resulting in multiple refusals to answer, thereby diminishing effective exploratory steps. Overall, in the majority of cases, PreAct outperforms ReAct. Furthermore, the application of Reflexion on top of PreAct consistently elevates model performance. 
Table \ref{tab:tot} presents the performance of PreAct and ReAct using the TOT selection strategy. Both 100-sample and 1000-sample results show that PreAct is approximately 5\% higher than ReAct, indicating a certain level of independence between PreAct and selection strategies like TOT.
These results suggest that the improvements in planning and decision-making ability in LLMs can be jointly provided by rich prior task information and observation predictions.

\subsection{Historical Prediction Influence Scope(RQ2)}
\label{history}

\begin{figure}[tbp]
	\centering
	\subfloat[HH]{\includegraphics[width=.45\columnwidth]{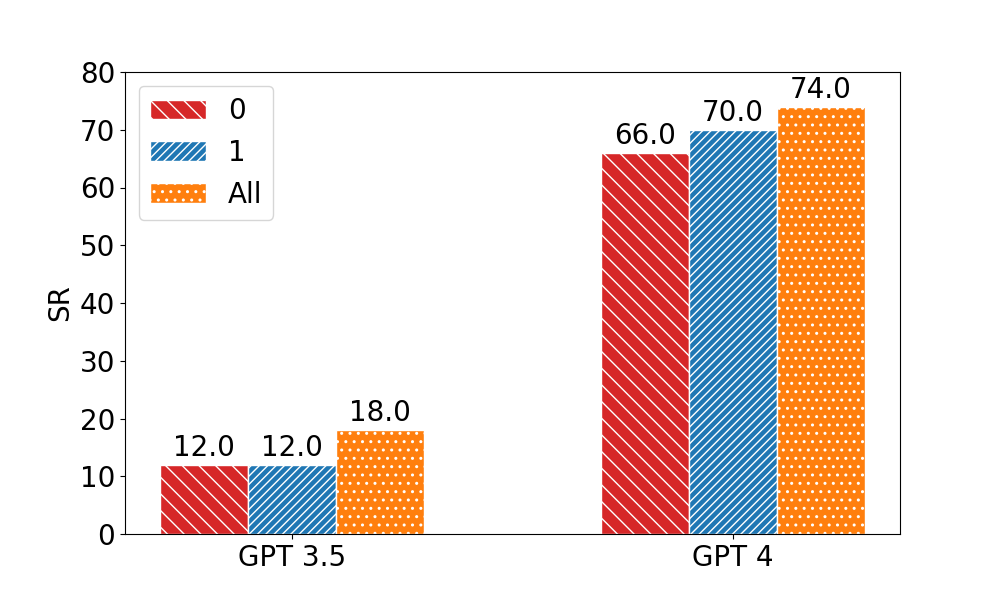}}\hspace{5pt}
	\subfloat[OS]{\includegraphics[width=.45\columnwidth]{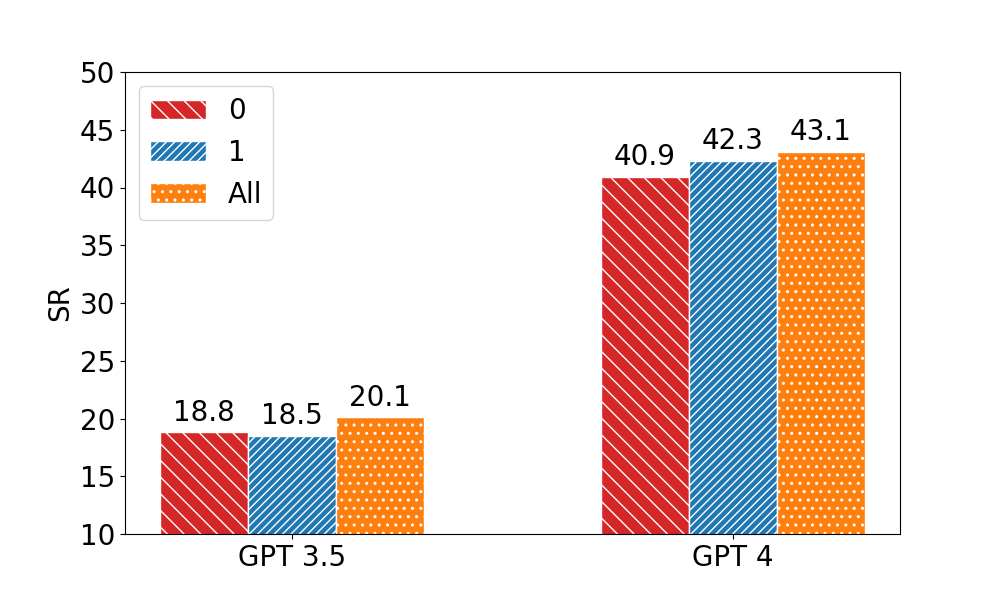}}\\
	\subfloat[DB]{\includegraphics[width=.45\columnwidth]{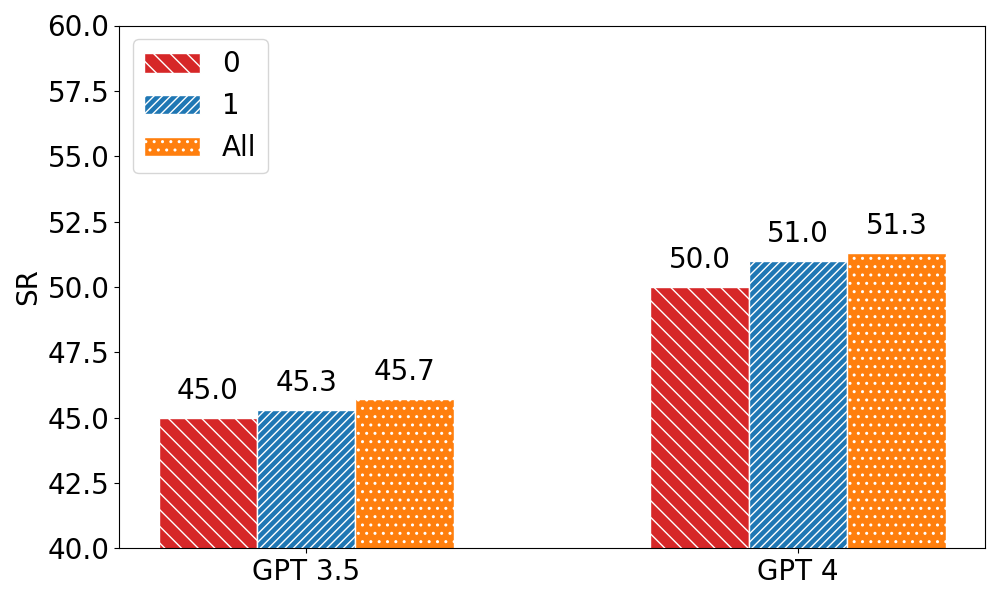}}\hspace{5pt}
	\subfloat[LTP]{\includegraphics[width=.45\columnwidth]{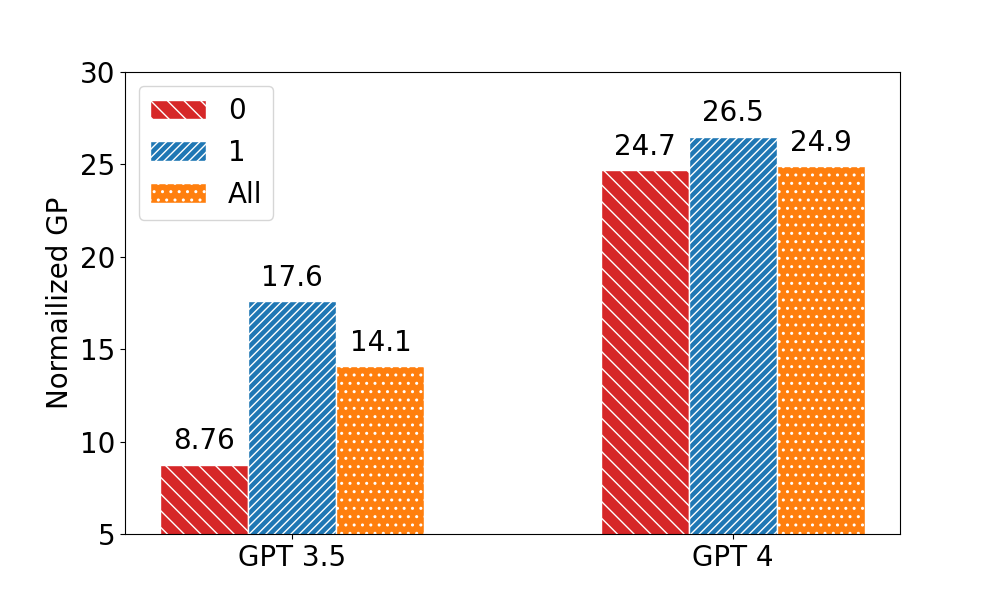}}
	\caption{Historical Prediction's Influence. 0 refers to ReAct's history, 1 refers to immediate mode history and all refers to permanent mode history.}
 \label{fig: historical}
\end{figure}

Figure \ref{fig: historical} demonstrates the impact of varying amounts of prediction history on the inferential performance of LLMs.\footnote{Zero historical prediction setting in this section is not equivalent to react in Section \ref{main result}, as the PreAct prompt is used to initiate the LLM in these experiments.}
It is evident from the experiments conducted on the HH, OS, and DB datasets that increased retention of prediction history correlates with a higher success rate. Take PreAct (GPT4) as an example, the success rate of tasks in 3 settings are 66\%, 70\%, 74\% in HH; 40.9\%, 42.3\%, 43.1\% in OS; and 50\%, 51\%, 51.3\% in DB, respectively. These findings suggest that historical predictions have a sustained positive effect on the model's reasoning abilities. However, on the LTP dataset, a greater amount of historical data results in a higher refusal probability, which in turn leads to a decline in performance in Permanent mode.


\begin{figure*}[t]
  \centering
  \includegraphics[width=0.95 \textwidth]{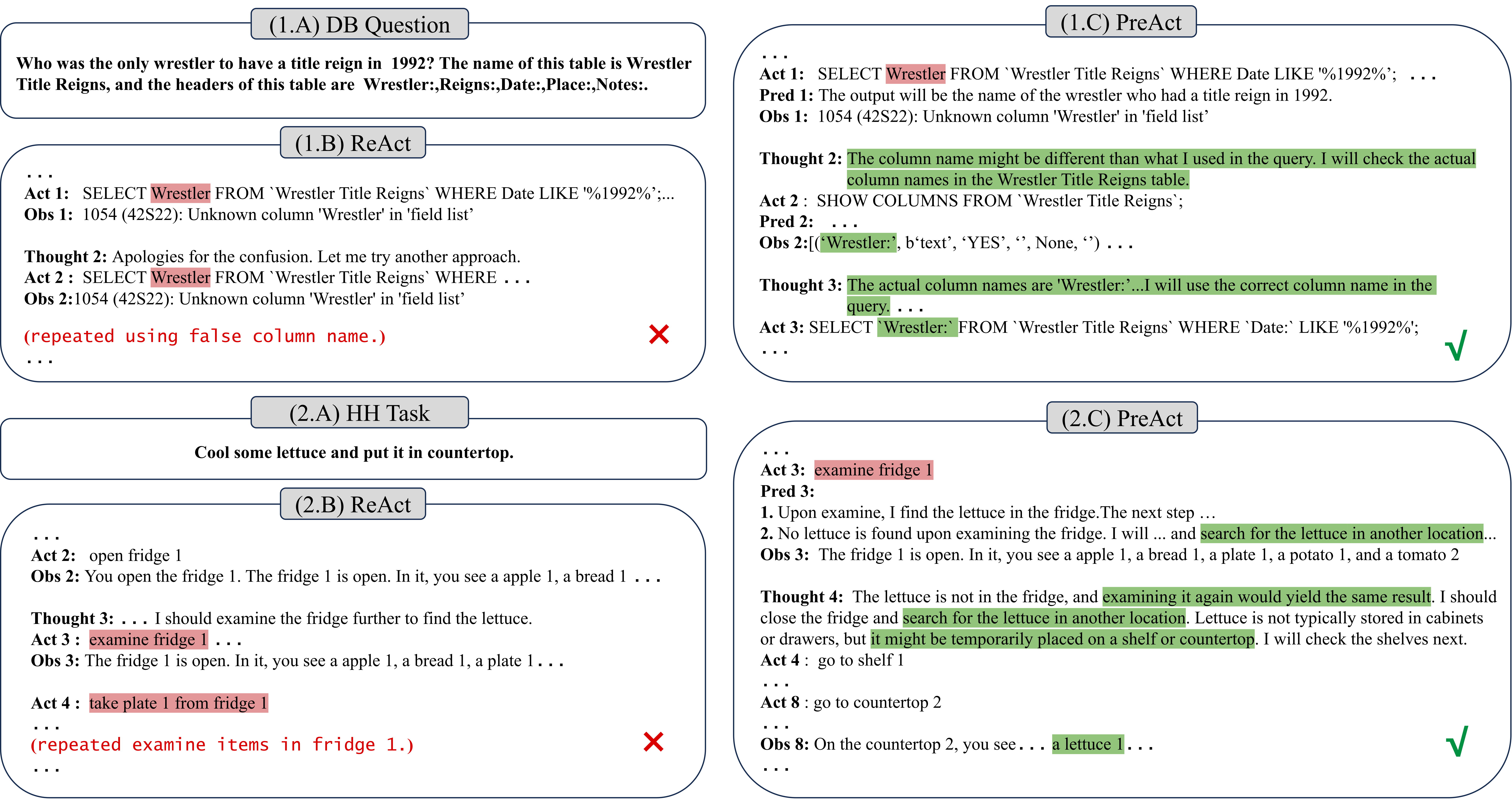}
  \caption{Two representative examples in DB and HH set between ReAct and PreAct. We omit unimportant information in the example. Act=Action, Obs=observation, Pred=prediction.}
  \label{fig:example_PreAct}
\end{figure*}

\begin{figure}[tbp]
    \centering
    \includegraphics[width=0.90\linewidth]{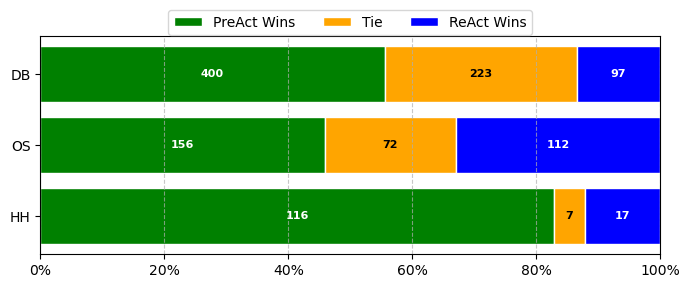}
    \caption{Overall Diversity Comparison between ReAct and PreAct}
    \label{fig:all-diversity}
\end{figure}

\subsection{Intrinsic Reason Analysis (RQ3)}
In our hypothesis, PreAct is presumed to enhance the inferential diversity and the directional strategy of reasoning, thereby augmenting the planning capabilities of the LLM. In this section, we will investigate these two contributing factors.

\textbf{Diversity}  Figure \ref{fig:all-diversity} displays the diversity comparison between ReAct and PreAct. The chart reveals that on any given dataset, at least 45\% of the instances show that PreAct thought's diversity is superior to ReAct, while the opposite scenario does not exceed 34\%. This indicates that using PreAct can significantly increase reasoning diversity, thereby expanding the inferential space and broadening the spectrum of possible actions. \footnote{Detailed information can be seen in Appendix \ref{diveristy}.}



\begin{table}[tbp]
  \centering
    \resizebox{0.35\textwidth}{!}{
    \begin{tabular}{lcccc}
    \Xhline{1.2pt}
    \multicolumn{1}{c}{\multirow{1}[4]{*}{Model}} & \multicolumn{2}{c}{Dev} & \multicolumn{2}{c}{Test} \\
    \cline{2-5}      & GPT3.5 & GPT4  & GPT3.5 & GPT4 \\
    \Xhline{1.2pt}
    ReAct & 0.69  & 1.89  & 0.85  & 1.91\\
    PreAct & \textbf{0.84}  & \textbf{2.29}  & \textbf{1.04}  & \textbf{2.30}\\
    \Xhline{1.2pt}
    \end{tabular}%

    }
  \caption{The score of strategy in HH dataset }
      \label{tab:strategy}%
\end{table}%

\textbf{Directional Strategy} 
As shown in Table \ref{tab:strategy}, PreAct's directional strategy score\footnote{We choose the Alfworld task to analyze the directional strategy, each trajectory will be scored $-1\sim3$ Appendix \ref{strategy} shows the score rules and its faithfulness.} is at least 20\% higher than that of ReAct. This indicates that PreAct is better at determining planning direction.

\textbf{Case Study} Figure \ref{fig:example_PreAct} shows the partial trajectories of PreAct and ReAct on the DB and HH datasets. Although PreAct and ReAct made identical errors, PreAct can rectify its mistakes, while ReAct does not. In the DB set, both ReAct and PreAct used the same incorrect column name in Act 1. PreAct corrected this by verifying the actual column names, while ReAct repeatedly used the erroneous column name. 
In the HH task, after examining the fridge, ReAct interacts with objects inside the fridge, which is irrelevant to the task, whereas PreAct had predicted $"$ No lettuce in fridge $"$ condition and tried to locate the lettuce elsewhere. 


\section{Conclusion}
In this paper, we introduce PreAct, a simple framework that utilizes predictions to enhance the diversity and strategic direction of planning, thereby improving the effectiveness of agents. This enhancement is continuous, independent of Reflexion, or TOT, and will persistently improve with the accumulation of historical predictions. 

Based on the findings of PreAct, we propose two metrics for evaluating planning, which may help in setting the reward functions at the process level for reinforcement learning in future work, ultimately training more powerful agents.


\section*{Limitations}
While PreAct improves the agent's planning ability, there are still directions to explore for future work.
(1) In most of the time, PreAct only interacts with the short-time memory like history. In the future, we will investigate the interaction between PreAct and other long-term memory beyond Reflexion.

(2) We only explore PreAct's ability by prompting, in the future, we will fine-tune the model with PreAct trajectory to find more intrinsic reason.

\section*{Broader Impact}
PreAct proposes that a model's reasoning and planning abilities can be enhanced through predictions which \textbf{presents a new approach for the implementation of LLM agents}. We have demonstrated that PreAct contributes to the improvement of reasoning diversity and directional strategic behavior, providing reasonable evaluation metrics for the reasoning of LLM agents. This will \textbf{have a positive impact on the assessment and optimization of LLM agents}.

However, due to the inherent hallucinations and biases of LLMs, PreAct may still exhibit deviations in intent, although it is known from LTP examples that PreAct, compared to Act-only models, possesses a stronger ability to refuse responses when faced with toxic texts.

\bibliography{custom}

\appendix
\section{Experiment Setup}
\label{exp-set}
\subsection{Hyper-parameter}
We use gpt-3.5-turbo-1106 in all 3.5 versions and gpt-4-1106-preview in all 4 versions in AgentBench.
We use gpt-4-turbo-2024-04-09 in TOT

\subsection{Dataset Information}
There are 4 datasets in AgentBench: Householding, Operating System, Database, Lateral Thinking Puzzles. \footnote{AgentBench uses the \textbf{original} Alfworld environment and MySQL environment. The other 2 datasets are created by AgentBench themselves. The difference between AgentBench and the original paper is just the format of the prompt, and they are both ReAct-based prompts. }

The Householding task (HH) uses the Alfworld benchmark \citep{shridhar2020alfworld}.
The ALFWorld benchmark consists of text-based simulations of home settings, offering an interactive platform for an agent to execute decision-making tasks via text interfaces. The agent's goal is to decompose a complex goal into simple actions, based on the provided environment description and a target instruction. With each action, the agent gets feedback from the environment, enabling it to adjust its strategy and proceed with the next task to achieve the primary goal ultimately.

The Operating System dataset (OS) \citep{liu2023agentbench} is designed to test Large Language Models (LLMs) by having them interact with and control an operating system through a terminal interface. It aims to assess LLMs within authentic interactive bash environments, such as Ubuntu Docker, by asking them questions with definitive answers or instructing them to perform a sequence of practical operations.

The Database dataset (DB) amalgamates several established datasets: WikiSQL \citep{zhong2017seq2sql}, WikiTableQuestions \citep{pasupat2015compositional}, SQA \citep{iyyer2017search}, HybridaQA \citep{chen2020hybridqa}, and FeTaQA \citep{nan2022fetaqa}, focusing on evaluate LLMs on authentic SQL interfaces and databases as is in the real world.

\label{ltp info}
The Lateral thinking puzzles (LTP) \citep{liu2023agentbench}, also known as situation puzzles are a globally popular group game. In the game, one person acts as the host and presents a puzzle, while other players attempt to solve it by asking questions that can only be answered with "yes," "no," or "irrelevant." The game ends when a player figures out the key aspect of the puzzle's story. The name of the game comes from the term "lateral thinking," which is the capacity to reason and generate new ideas from unique and unconventional viewpoints.

Since some LTP's contents and descriptions might induce psychological discomfort, the application of ReAct and PreAct in LTP may trigger GPT's rejection mechanism leading fail. As a result, the original LTP dataset uses the Act-only framework. For a fair comparison, we remove all samples that have refusal for more than three (included) consecutive rounds and change the metric of LTP to a normalized one.

More detailed information about these 4 datasets has been shown below.
\begin{table}[htbp]
  \centering
    \resizebox{0.5\textwidth}{!}{
    \begin{tabular}{lcccc}
    \Xhline{1.5pt}
    Dataset & Operating System & Database & Householding & Lateral Thinking Puzzles \\
    \Xhline{1.5pt}
    Abbr. & OS    & DB    & HH    & LTP\\
    \#Max Turn & 8     & 5     & 35    & 50 \\
    Metrics & SR    & SR    & SR    & Normalized GP \\
    \#Dev  & 26    & 60    & 20    & 40 \\
    \#Test & 144   & 300   & 50    & 100\\
    \Xhline{1.5pt}
    \end{tabular}%
    }
  \caption{The information of 4 datasets in AgentBench\citep{liu2023agentbench}. In the main text, we use abbreviations to represent these datasets. "\#Max Turn" denotes the estimated number of interacting max turns to solve a single problem.  "SR" stands for Success Rate. "Normalized GP" refers to the percent of information the agent has guessed in the games which will not lead to refusal for more than three consecutive rounds. "\#Dev" is the number of samples in the Dev set. "\#Test" is the number of samples in the Test set. We align with the setting in AgentBench's GitHub code, which may be different from the setting in the paper.}
      \label{tab:dataset info}%
\end{table}%

In the rounds following the successful inference of "The tools pierced through Xiao Ming's skull", there will be safety prompts that output in an unexpected format, which leads to the failure of PreAct. (It should be noted that this is not the input exceed error, since they have different error logs!)

\begin{tcolorbox}[
colback=white!10!white,
colframe=PastelPurple,
title=Harmful example in ltp,
breakable]

Question: The classmates celebrated Xiao Ming's birthday, but under the watchful eyes of everyone, Xiao Ming was stabbed to death by a sharp weapon, please reason.
\\
\\
Ground Truth: Some cakes need to be shaped, and they will be fixed with bamboo sticks or similar sharp objects. In order to surprise Xiao Ming, the classmate who ordered the cake ordered a cake with a strange shape, which contained a weapon. However, some students didn't know that after singing the birthday song to Xiao Ming, they put Xiao Ming's head into the cake and pierced his head with a sharp weapon.
\end{tcolorbox}

We use HotpotQA \citep{yang2018hotpotqa} in TOT setting. The TOT-ReAct version comes from LATS \citep{zhou2023language}. Due to the inherent randomness of TOT results and the large scale of HotpotQA, we ran the test 5 times with a sample size of 100 (id 0-99) and 1 time with a sample size of 1000 (id 0-999). 

\section{Prompts}
\label{PreAct prompt}
Italic text will be replaced by real task information or trajectory.

\subsection{Prompts in HH}
\begin{tcolorbox}[
colback=white!10!white,
colframe=PastelPurple,
title=HH PreAct Prompt,
breakable]
Interact with a household to solve a task. Imagine you are an intelligent agent in a household environment and your target is to perform actions to complete the task goal. At the beginning of your interactions, you will be given the detailed description of the current environment and your goal to accomplish. For each of your turn, you will be given a list of actions which you can choose one to perform in this turn. In each of your turn, you must first think about the current condition and plan for your future actions, and then output your action in this turn, and then predict the various types of feedback the environment might provide at a high level, and ensure to furnish corresponding handling measures for each potential category of feedback. Your output must strictly follow this format:"THOUGHT: your thoughts.\\
ACTION: your next action.\\
PREDICTED FEEDBACK:\\
1. first possible feedback type and the corresponding handling measures.\\
2. second possible feedback type and the corresponding handling measures.\\
and so on...\\
". After your each turn, the environment will give you immediate feedback based on which you plan your next few steps. if the environment output "Nothing happened", that means the previous action is invalid and you should try more options. If the actual feedback doesn't fall into any of the previously predicted scenarios, you need to contemplate the reasons for this disparity in the next round's "THOUGHT:" section and use this as a basis to reflect on the previous planning. Finally, integrate all relevant factors to re-plan the strategy.\\
Reminder: \\
1. The action must be chosen from the given available actions. Any actions except provided available actions will be regarded as illegal.\\
2. The records of available actions from past rounds will be deleted, and only the available actions for the current round will be provided. You need to adapt to this reduction and correctly understand the context.\\
3. In each of your turn, you must first think based on the task, current environment, and historical information before providing an action, then predict the types of environmental feedback. You cannot skip any of these steps and your output must strictly follow the format provided before.\\

\end{tcolorbox}

\begin{tcolorbox}[
colback=white!10!white,
colframe=PastelPurple,
title=HH ReAct Prompt,
breakable]
Interact with a household to solve a task. Imagine you are an intelligent agent in a household environment and your target is to perform actions to complete the task goal. At the beginning of your interactions, you will be given the detailed description of the current environment and your goal to accomplish. For each of your turn, you will be given a list of actions which you can choose one to perform in this turn. You should choose from two actions: "THOUGHT" or "ACTION". If you choose "THOUGHT", you should first think about the current condition and plan for your future actions, and then output your action in this turn. Your output must strictly follow this format:"THOUGHT: your thoughts.\\ ACTION: your next action\\"; If you choose "ACTION", you should directly output the action in this turn. Your output must strictly follow this format:"ACTION: your next action\\". After your each turn, the environment will give you immediate feedback based on which you plan your next few steps. if the environment output "Nothing happened", that means the previous action is invalid and you should try more options.\\ Reminder: \\1. the action must be chosen from the given available actions. Any actions except provided available actions will be regarded as illegal. \\2. Think when necessary, try to act directly more in the process.\\\\"

\end{tcolorbox}

\subsection{Prompts in OS}
\begin{tcolorbox}[
colback=white!10!white,
colframe=PastelPurple,
title=OS PreAct Prompt,
breakable]
You are an assistant that will act like a person, I'will play the role of linux(ubuntu) operating system. Your goal is to implement the operations required by me or answer to the question proposed by me. For each of your turn, you should first think what you should do, then you should take exact one of the three actions: "bash", "finish" or "answer". If your action is 'bash', you should also predict the possibile OS output in HIGH LEVEL and give suggestions for each possibility to deal with it. You should also compare the real output of the OS, your last prediction and the suggestion. If the real output mismatch your prediction, you should reflect your thought and action and make new planning.

1. If you think you should execute some bash code, take bash action, and you should print like this:

Think: put your thought here.

Act: bash

\texttt{```}
bash

\# put your bash code here

\texttt{```}

Predict OS output:\\
1.Your first prediction and suggestion\\
2.Your second prediction and suggestion\\
...\\

2. If you think you have finished the task, take finish action, and you should print like this:

Think: put your thought here.

Act: finish

3. If you think you have got the answer to the question, take answer action, and you should print like this:

Think: put your thought here.

Act: answer(Your answer to the question should be put in this pair of parentheses)

If the output is too long, I will truncate it. The truncated output is not complete. You have to deal with the truncating problem by yourself. Attention, your bash code should not contain any input operation. Once again, you should take only exact one of the three actions in each turn.

\textit{(example)}

Now, I will start a new problem in a new OS. My problem is:

\textit{(problem)}

\end{tcolorbox}

\begin{tcolorbox}[
colback=white!10!white,
colframe=PastelPurple,
title=OS ReAct Prompt,
breakable]

You are an assistant that will act like a person, I'will play the role of linux(ubuntu) operating system. Your goal is to implement the operations required by me or answer to the question proposed by me. For each of your turn, you should first think what you should do, and then take exact one of the three actions: "bash", "finish" or "answer".

1. If you think you should execute some bash code, take bash action, and you should print like this:

Think: put your thought here.

Act: bash

\texttt{```}
bash

\# put your bash code here

\texttt{```}

2. If you think you have finished the task, take finish action, and you should print like this:

Think: put your thought here.

Act: finish

3. If you think you have got the answer to the question, take answer action, and you should print like this:

Think: put your thought here.

Act: answer(Your answer to the question should be put in this pair of parentheses)

If the output is too long, I will truncate it. The truncated output is not complete. You have to deal with the truncating problem by yourself. Attention, your bash code should not contain any input operation. Once again, you should take only exact one of the three actions in each turn.

\textit{(example)}

Now, I will start a new problem in a new OS. My problem is:

\textit{(problem)}

\end{tcolorbox}

\subsection{Prompts in DB}
\begin{tcolorbox}[
colback=white!10!white,
colframe=PastelPurple,
title=DB PreAct Prompt,
breakable]
I will ask you a question, then you should help me operate a MySQL database with SQL to answer the question.\\
You should thought, give act, and predicte the possible output of the SQL.\\
In thought part, you should explain the problem and your solution to me. If the SQL output mismatches the predict output, you should ckeck the plan and SQL in the last round carefully, find out its mistake in it and refine it.\\
After thinking and explaining thoroughly, every round you can choose to operate or to answer. 
Finally, you should predict the possible HIGH LEVEL output of the SQL and give a next step suggestion. Remember your prediction should be HIGH LEVEL, not just the SQL output.\\
your operation should be like this:

Thought: Your thoughts here\\
Action: Operation\\
\texttt{```}sql\\
SELECT * FROM table WHERE condition;\\
\texttt{```}\\
Predict MySql Output:\\
1. Your first HIGH LEVEL output of the SQL and the next step suggestion.\\
2. Your second HIGH LEVEL output of the SQL and the next step suggestion.\\
...

You MUST put SQL in markdown format without any other comments. Your SQL should be in one line.\\
Every time you can only execute one SQL statement. I will only execute the statement in the first SQL code block. Every time you write a SQL, I will execute it for you and give you the output.\\
If you are done operating, and you want to commit your final answer, then write down:

Thought: Your thoughts here\\
Action: Answer\\
Final Answer: ["ANSWER1", "ANSWER2", ...]\\

DO NOT write this pattern unless you are sure about your answer. I expect an accurate and correct answer.\\
Your answer should be accurate. Your answer must be exactly the same as the correct answer.\\
If the question is about modifying the database, then after done operation, your answer field can be anything.\\
If your response cannot match any pattern I mentioned earlier, you will be judged as FAIL immediately.\\
Your input will be raw MySQL response, you have to deal with it by yourself.
\end{tcolorbox}

\begin{tcolorbox}[
colback=white!10!white,
colframe=PastelPurple,
title=DB ReAct Prompt,
breakable]
I will ask you a question, then you should help me operate a MySQL database with SQL to answer the question.\\
You have to explain the problem and your solution to me and write down your thoughts.\\
After thinking and explaining thoroughly, every round you can choose to operate or to answer.\\
your operation should be like this:\\

Thought: Your thoughts here\\
Action: Operation\\
\texttt{```}sql\\
SELECT * FROM table WHERE condition;\\
\texttt{```}\\

You MUST put SQL in markdown format without any other comments. Your SQL should be in one line.\\
Every time you can only execute one SQL statement. I will only execute the statement in the first SQL code block. Every time you write a SQL, I will execute it for you and give you the output.\\
If you are done operating, and you want to commit your final answer, then write down:\\
Action: Answer\\
Final Answer: ["ANSWER1", "ANSWER2", ...]\\
DO NOT write this pattern unless you are sure about your answer. I expect an accurate and correct answer.\\
Your answer should be accurate. Your answer must be exactly the same as the correct answer.\\
If the question is about modifying the database, then after done operation, your answer field can be anything.\\
If your response cannot match any pattern I mentioned earlier, you will be judged as FAIL immediately.\\
Your input will be raw MySQL response, you have to deal with it by yourself.\\
\end{tcolorbox}

\subsection{Prompts in LTP}
Following is the prompt of LTP. The Chinese version is just a translation of the English version.

\begin{tcolorbox}[
colback=white!10!white,
colframe=PastelPurple,
title=LTP PreAct Prompt,
breakable]
You are a game player, and you are playing Lateral Thinking Puzzle, also known as Situation Puzzle.\\
Lateral Thinking Puzzle is a deductive reasoning game, and here are the game rules:\\
1. At the beginning of the game, you will receive a narrative, referred to as "story". Based on the story, you need to ask questions that can be answered with "yes", "no", or "irrelevant" to guess out the "truth".\\
2. By asking questions, you narrow down the range of possibilities until you eventually guess out the truth.\\
3. Each time, you can only ask one question.\\
4. Remember that your role is a player. You cannot declare the end of the game, give up on reasoning, or request a new game.\\
5. You cannot directly repeat information already provided in the story.\\
6. You cannot directly ask for details about the story in the form of "why" questions; you need to make your own guesses for truth.\\
7. You cannot directly inquire about the story; you must make your own deductions.\\

Next, please make full use of the information provided above to engage in game reasoning. Keep in mind that your questions should be answerable with "yes", "no", or "irrelevant", and you can only ask one question at a time.\\
In order for you to deduce the truth from the story more effectively, in each of your turn, you must output your question, and then give a plan of question direction for each potential category of feedback the host might provide, which is among Yes, No, Irrelevant and Redundant. In each of your turn, please ensure that your output contains a question about the story and strictly adhere to the following template: "Question: [Your question in this turn].\\
Predicted Feedback:\\
1.	Yes. [Next turn plan].\\
2.	No. [Next turn plan].\\
3.	Irrelevant. [Next turn plan].\\
4.	Redundant. [Next turn plan]."\\
Note that we may delete some Predicted Feedback in the history, but you should follow the Question, Predicted Feedback format.\\
Here is your story:\\
\textit{(story)}\\

You can start guessing the content of the truth, and I will answer your questions. Please note that your questions should be answerable with "yes", "no", or "irrelevant".

\end{tcolorbox}

\begin{tcolorbox}[
colback=white!10!white,
colframe=PastelPurple,
title=LTP ReAct Prompt,
breakable]
You are a game player, and you are playing Lateral Thinking Puzzle, also known as Situation Puzzle.\\
Lateral Thinking Puzzle is a deductive reasoning game, and here are the game rules:\\
1. At the beginning of the game, you will receive a narrative, referred to as "story". Based on the story, you need to ask questions that can be answered with "yes", "no", or "irrelevant" to guess out the "truth".\\
2. By asking questions, you narrow down the range of possibilities until you eventually guess out the truth.\\
3. Each time, you can only ask one question.\\
4. Remember that your role is a player. You cannot declare the end of the game, give up on reasoning, or request a new game.\\
5. You cannot directly repeat information already provided in the story.\\
6. You cannot directly ask for details about the story in the form of "why" questions; you need to make your own guesses for truth.\\
7. You cannot directly inquire about the story; you must make your own deductions.\\

Next, please make full use of the information provided above to engage in game reasoning. Keep in mind that your questions should be answerable with "yes", "no", or "irrelevant", and you can only ask one question at a time.\\
Here is your story:\\
\textit{(story)}\\

You can start guessing the content of the truth, and I will answer your questions. Please note that your questions should be answerable with "yes", "no", or "irrelevant".

\end{tcolorbox}

\subsection{Prompt of Reflexion}
Both ReAct and PreAct use the same format, we just set examples to '' in Preact.
\begin{tcolorbox}[
colback=white!10!white,
colframe=PastelPurple,
title=Reflexion Prompt,
breakable]
You are an advanced reasoning agent that can improve based on self reflection. You will be given a previous reasoning trial and a question to answer. You were unsuccessful in answering the question either because you guessed the wrong answer, or you used up your set number of reasoning steps. In a few sentences, Diagnose a possible reason for failure and devise a new, concise, high level plan that aims to mitigate the same failure. Use complete sentences.  \\
Here are some examples: \\
\textit{\{examples\}}\\
(END OF EXAMPLES)\\

Previous trial:\\
Question: \textit{\{question\}}\\

\textit{\{scratchpad\}}\\

Reflection:
\end{tcolorbox}

\subsection{Prompt of PreAct in HotpotQA}
\begin{tcolorbox}[
colback=white!10!white,
colframe=PastelPurple,
title=HH ReAct Prompt,
breakable]
Solve a question answering task with interleaving Thought, Action, Predicted Feedback, Observation steps. Thought can reason about the current situation, and Action can be three types:\\ 
(1) Search[entity], which searches the exact entity on Wikipedia and returns the first paragraph if it exists. If not, it will return some similar entities to search.\\
(2) Lookup[keyword], which returns the next sentence containing keyword in the current passage.\\
(3) Finish[answer], which returns the answer and finishes the task.\\
Predicted Feedback are your guess at the possible feedback type and the corresponding handling measures before observation. \\
Thoughts, Actions, and Predicted Feedback should be one line each, so du not use multiple lines inside them.\\
After each observation, provide the next Thought and next Action. Here are some examples:\\
\textit{(three examples)}
\\
\\
\textit{(problem)}
\end{tcolorbox}

\subsection{Prompt of Diversity Judgment}
\begin{tcolorbox}[
colback=white!10!white,
colframe=PastelPurple,
title=Directional Strategy Judgment Prompt,
breakable]
\label{diveristy prompt}
I will provide you two trajectories of an agent interacting with the environment to accomplish the same task. Please evaluate the diversity of the agent's thinking and actions in these two trajectories and assign a score (0 to 100) for each trajectory. When evaluating, please analyze and compare the given trajectories first, provide your thought process, and then give the final diversity score for each trajectory. Your output should strictly adhere to the following format:\\
"Thought: [Your Thought]\\
Score 1: [Score of trajectory 1]\\
Score 2: [Score of trajectory 2]"\\

[BEGIN OF ONE TRAJECTORY]

\textit{(one of the trajectories) }

[END OF ONE TRAJECTORY]

[BEGIN OF ANOTHER TRAJECTORY]

\textit{(another of the trajectories) }

[END OF ANOTHER TRAJECTORY]
\end{tcolorbox}

\subsection{Prompt of Directional Strategy Judgment}
Following is the prompt of directional strategy judgment. We remove all predictions in the history and turn for a fair comparison The red part is the criteria of scoring.
\begin{tcolorbox}[
colback=white!10!white,
colframe=PastelPurple,
title=Directional Strategy Judgment Prompt,
breakable]
\label{directional strategy prompt}
I will provide you with a part of the trajectory where an agent interacts with the environment to accomplish a certain task, and the complete ground truth trajectory of the task. You need to assess the quality of the direction of the action plan in the last round of the evaluated trajectory at a high level, that is, what extent it facilitated the completion of the task with the information agent has gained, and provide a score (minus one to positive three) for the last round of the trajectory. \textcolor{red}{(3: last plan direction is correct based on ground truth, 2: last plan direction is incorrect based on ground truth but seems (strong) reasonable based on evaluated trajectory history, 1: last plan direction is incorrect based on ground truth but seems (weak) reasonable based on evaluated trajectory history, 0: last plan direction is incorrect based on ground truth and seems unreasonable based on evaluated trajectory history, and the direction of evaluated trajectory history is also incorrect, -1: last plan direction is incorrect based on ground truth but the direction of evaluated trajectory history is correct, last plan direction disturb the direction)}

You must first analyze and understand the reasons for the success of the ground truth trajectory, and then analyze what agent know in the evaluated trajectory, and then analyze the impact of changing the action plan on completing the task and measure the extent to which these effects facilitate task completion with a score. \\
Your output should strictly adhere to the following format: \\
Thought: [Your Thought] \\
Last Round Replan Score: [Score for the last round replan] \\

[BEGIN OF GROUND TRUTH TRAJECTORY] 

\textit{(ground truth trajectory) }

[END OF GROUND TRUTH TRAJECTORY] 

[BEGIN OF THE TRAJECTORY TO BE EVALUATED] 

\textit{(evaluated trajectory history)}

\textit{(evaluated trajectory turn)}

[END OF THE TRAJECTORY TO BE EVALUATED] 
\end{tcolorbox}

\section{Correlation Analysis}
\label{Correlation}
\begin{figure}[htbp]
    \vspace{-5mm}
	\centering
	\subfloat[Diversity Correlation in Dev set]{\includegraphics[width=.9\columnwidth]{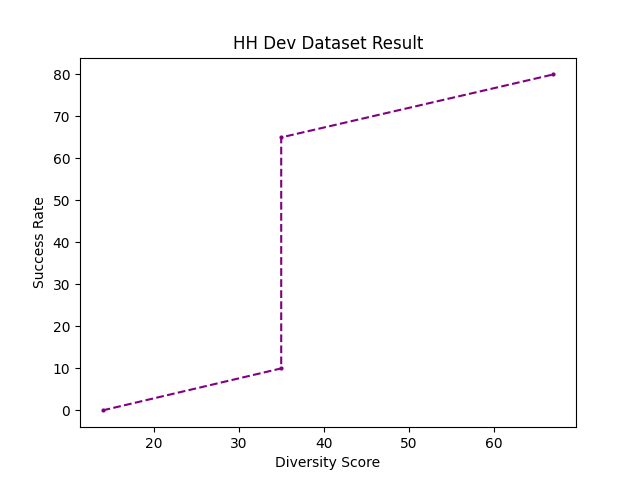}}\hspace{1pt}
	\subfloat[Diversity Correlation in Test set]{\includegraphics[width=.9\columnwidth]{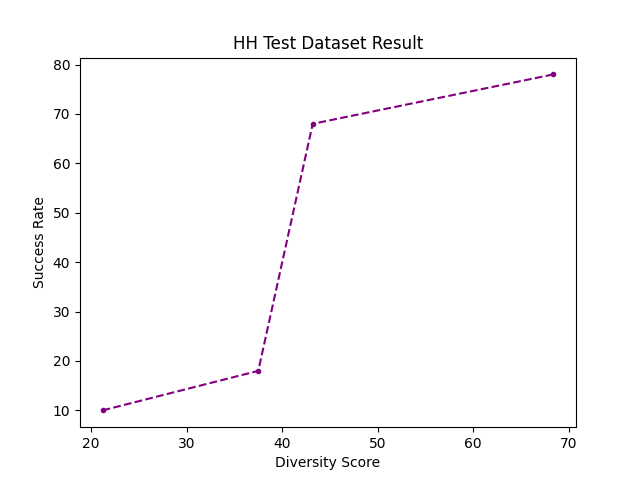}}\\
	\subfloat[Directional Strategy Correlation in Dev set with GPT4]{\includegraphics[width=.9\columnwidth]{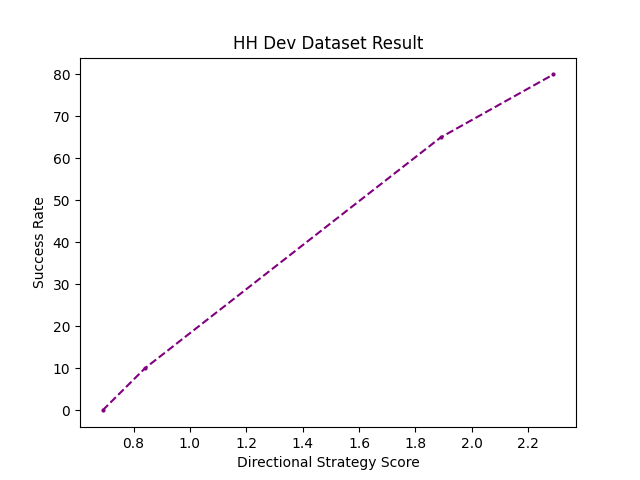}}\hspace{1pt}
	\subfloat[Directional Strategy Correlation in Test set]{\includegraphics[width=.9\columnwidth]{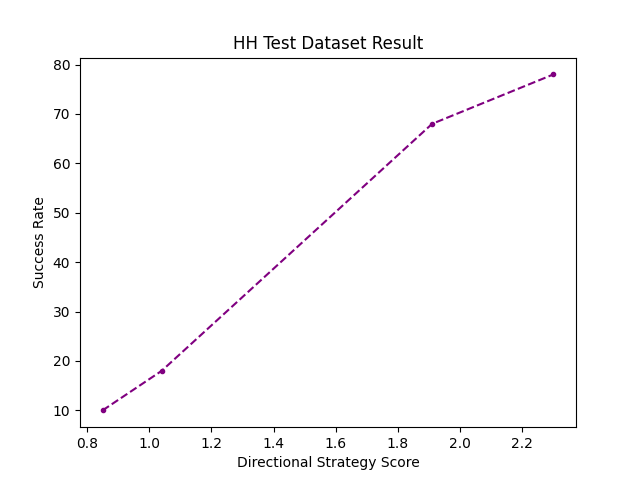}}
	\caption{Correlation Analysis in HH Dataset}
 \label{fig: Correlation}
\end{figure}

Figure \ref{fig: Correlation} displayed the relationship between diversity, directional strategy, and success rate on HH, revealing that the success rate is positively correlated with both indicators. Furthermore, the correlation coefficient between directional strategy and success rate is 99.8\% (Dev) and 99.3\% (Test), whereas the correlation coefficient for diversity is 83.7\% (Dev) and 91.2\% (Test).

To further prove the reliability of the diversity metric, We computed the number of different actions (divided by the total round number) in Alfworld and found about 82\% of results can match our diversity result. This can partly prove the efficiency of Diversity metrics.

Since there are a lot of works use GPT to select the possible state (for example, TOT \citep{yao2023tree}, LATS \citep{zhou2023language}), it has been proved that GPT4 can judge directional strategy.

\section{Specific Diversity Comparison between ReAct and PreAct}
\label{diveristy}
We presented two trajectories with thought and action to GPT-4, asking it to score each trajectory on a scale from 0 to 100.\footnote{The prompt can be found in Appendix \ref{diveristy prompt}}
Figure \ref{fig: diversity specific} presents the specific diversity comparison between ReAct and PreAct among different set and model.
 
\section{Directional Strategy metrics.}
\label{strategy}
 For each round of every trajectory, we provide the model with ground truth, all thoughts and actions from previous rounds and the current round's thoughts and actions, while discarding all predictions. We then ask GPT-4 to score its directional strategy on a scale from $-1\sim3$ for each turn and the metric of strategy is: \begin{equation}
    M_{s} = E_{x\sim p}\left( E_{t\sim x}\left(LLM_{score}\left(t\right) \right)\right)
\end{equation}
where $x$ is the sample, $t$ is the thought and action in one turn and $LLM_{score}$ is the scorer.

\textbf{Score rules} 

3: last plan direction is correct based on ground truth

2: last plan direction is incorrect based on ground truth but seems (strong) reasonable based on evaluated trajectory history

1: last plan direction is incorrect based on ground truth but seems (weak) reasonable based on evaluated trajectory history

0: last plan direction is incorrect based on ground truth and seems unreasonable based on evaluated trajectory history, and the direction of evaluated trajectory history is also incorrect

-1: last plan direction is incorrect based on ground truth but the direction of evaluated trajectory history is correct, last plan direction disturb the direction.

\textbf{Faithfulness} To fairly compare whether the scores of PreAct and ReAct align with human expectations, we first selected 20 PreAct trajectories and chose one round from each as the annotation round. We then removed all predictions and used the ReAct method to obtain the corresponding annotation round for ReAct. For GPT4, we scored according to the scoring principle of prompting GPT-4. For human annotation, we provided the prompt and content given to GPT-4 to the annotators and emphasized the annotation principle. PreAct and ReAct trajectories were provided each time, but their order of appearance was random. We compared the Directional Strategy scores given by each annotator for these 40 decisions with the Directional Strategy scores given by GPT-4.
The experiment found that humans and GPT-4 completely agreed on 50\% of the data, and for 85\% of the data, the score difference between humans and GPT-4 was less than or equal to 1 point.

\begin{figure}[htbp]
	\centering
	\subfloat[Diversity in Dev set with GPT3.5]{\includegraphics[width=.95\columnwidth]{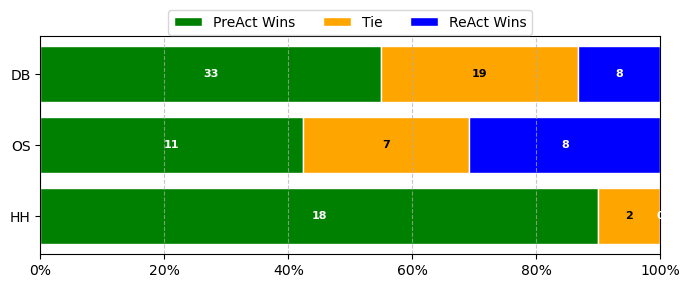}}\hspace{5pt}
	\subfloat[Diversity in Test set with GPT3.5]{\includegraphics[width=.95\columnwidth]{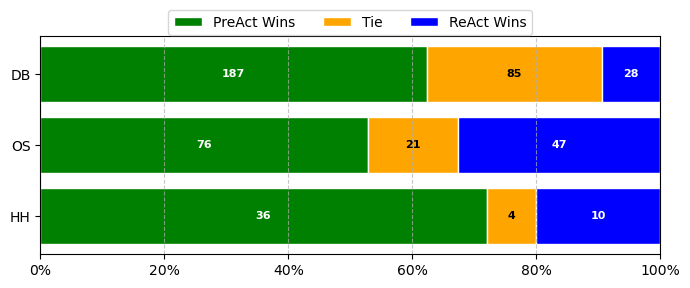}}\\
	\subfloat[Diversity in Dev set with GPT4]{\includegraphics[width=.95\columnwidth]{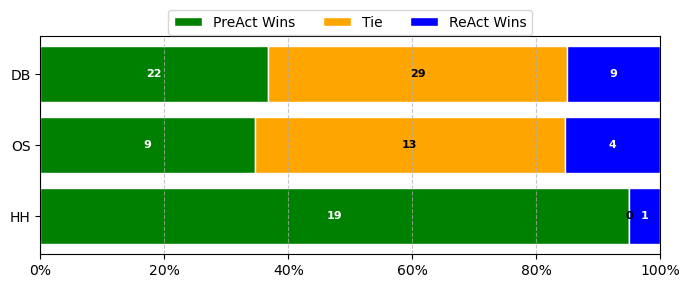}}\hspace{5pt}
	\subfloat[Diversity in Test set with GPT4]{\includegraphics[width=.95\columnwidth]{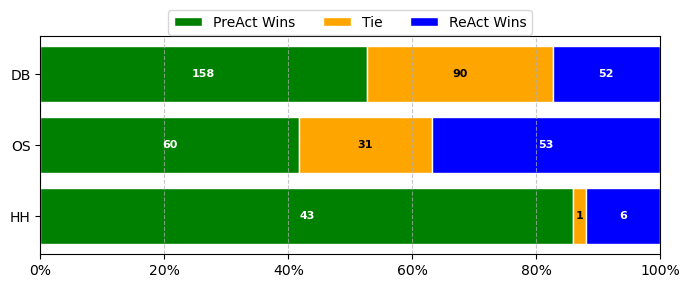}}
	\caption{Specific Diversity Comparison between ReAct and PreAct}
 \label{fig: diversity specific}
\end{figure}

\section{When PreAct Meets Hallucination}

Table \ref{tab:tot_h} displays the performance of PreAct+TOT under the condition of hallucination in prediction. Under this setup, we first run the PreAct+TOT setting with sample IDs 100-999 and save all predictions. During inference, the original prediction may be replaced with a random prediction from the corresponding round with the replacement rate being the value of hallucination (h in the Table \ref{tab:tot_h}.)

The result in Table \ref{tab:tot_h} shows that when all predictions are hallucinatory, the performance may decrease and the variance of the result is big, so PreAct may be largely influenced by hallucination. But when half of the predictions are hallucinatory, the performance can improve in most cases. As a result, if the degree of hallucination is not significant, PreAct can perform better.

\begin{table}[tbp]
  \centering
    \resizebox{0.45\textwidth}{!}{
    \begin{tabular}{lcccccc}
    \Xhline{1.2pt}
    Model & 0 & 1  & 2 & 3 & 4  \\
    \Xhline{1.2pt}
    ReAct+TOT & 66 & 63 & 62 & 65 & 67 \\
    PreAct+TOT & 70  & 72  & 67  & 70 & 68  \\
    PreAct+TOT (h=0.5) & 67 & 68 & 72 & 63 & 63 \\
    PreAct+TOT (h=1.0) & 66  & 61  & 69  & 55 & 69 \\
    \Xhline{1.2pt}
    \end{tabular}%
}
  \caption{The result of ReAct and PreAct with TOT in HotpotQA under different sample sizes. We ran the test 5 times with a sample size of 100.}
      \label{tab:tot_h}%
\end{table}%

\section{Related Work}
\subsection{Agent Planning}
With the discovery of the chain-of-thought \citep{wei2022chain,kojima2022large}, utilizing the reasoning capabilities of LLMs for planning has become possible \citep{huang2022language}. Within this context, two modes are distinguished: ReWOO \citep{xu2023rewoo} and ReAct \citep{yao2022react}.

When faced with a task, the former works \citep{xu2023rewoo,chen2022program,lu2023chameleon,hu2023chatdb} conducts all planning in one go and executes sequentially, while the latter executes planning step by step. Although ReWOO possesses higher efficiency and fewer model invocations, it struggles with complex, observation-requiring planning tasks. 

ReAct, on the other hand, synthesizes thought and action and continuously updates this approach based on observations, allowing it to cope with a wider variety of situations \citep{yao2022react,wang2023describe}. However, as this paper points out, ReAct's reasoning diversity and directional strategy are less robust. 

Works like Tree-of-Thought\citep{yao2023tree,hu2023tree} and Graph-of-thought\citep{besta2023graph,sun2023think} allow the generation of multiple possible actions at each step to expand the action space and explore the most likely directions. 

To more efficiently choose directions, the works like LLM-MCTS \citep{zhao2023large}, RAP \citep{hao2023reasoning}, LATS \citep{zhou2023language}, DoraemonGPT \citep{yang2024doraemongpt} and Toolchain \citep{zhuang2023toolchain}, employed pathfinding algorithms such as $A^*$ \citep{hart1968formal} or MCTS \citep{metropolis1949monte}.

\subsection{Agent long-term Memory}
In this paper, we only consider 1 type of agent long-term memory, Reflexion \citep{shinn2023reflexion}. Besides, there are still 2 types: example memory and insight memory.

Example memory entails the manual creation of samples that align with the expectations of specific tasks. During operation, instances of successful examples that are akin to the current task are retrieved using techniques such as vector similarity or BM25 \citep{dong2023demonsf,dong2023revisit,zhao-etal-2023-demosg}. These examples are then fed into the large language model as part of the prompt. \citep{wang2023voyager,wen2023dilu,song2023llm,zhong2023memorybank}

Conversely, insight memory encapsulates both successful and unsuccessful instances into condensed insights via the large language model. When faced with new tasks, these synthesized insights are incorporated directly into the prompt for the large language model, assisting in the planning and decision-making processes. \citep{majumder2023clin,zhao2023expel}.

\end{document}